\documentclass[11pt,a4paper]{article}

\usepackage[margin=0.8in]{geometry}
\usepackage{amsmath,amssymb,booktabs,graphicx,xcolor}
\usepackage{flafter}
\usepackage[hidelinks]{hyperref}

\newcommand{\system}{RegenHarness}
\makeatletter
\renewcommand\section{\@startsection{section}{1}{\z@}%
  {-2.8ex plus -0.8ex minus -.2ex}{1.3ex plus .2ex}%
  {\normalfont\large\bfseries\raggedright}}
\renewcommand\subsection{\@startsection{subsection}{2}{\z@}%
  {-2.2ex plus -.6ex minus -.2ex}{.9ex plus .2ex}%
  {\normalfont\normalsize\bfseries\raggedright}}
\renewcommand\paragraph{\@startsection{paragraph}{4}{\z@}%
  {1.5ex plus .5ex minus .2ex}{-.55em}%
  {\normalfont\normalsize\bfseries}}
\long\def\@makecaption#1#2{%
  \vskip 7pt
  {\small\noindent\textbf{#1.}\enspace #2\par}%
  \vskip 3pt}
\newenvironment{paperwidefigure}{%
  \begin{figure}[!htb]\centering
}{\end{figure}}
\newenvironment{paperanchoredfigure}{%
  \par\addvspace{10pt}\noindent\begin{minipage}{\textwidth}\def\@captype{figure}\centering
}{\end{minipage}\par\addvspace{10pt}}
\newenvironment{paperwidetable}{%
  \par\noindent\begin{minipage}{\textwidth}\def\@captype{table}\centering
}{\end{minipage}\par}
\makeatother

\title{\system: A Robot Agent Harness with\\
Evidence-Gated Recursive Self-Improvement}
\author{%
{\small\textbf{Kailin Wang}$^{1}$ \quad
\textbf{Haoxiang Jie}$^{1}$ \quad
\textbf{Yaoyuan Yan}$^{1}$ \quad
\textbf{Zhiyou Heng}$^{1}$ \quad
\textbf{Zhaosong Li}$^{2,3}$}\\[0.35em]
{\small $^{1}$AI Lab, Country Garden Services Group}\\
{\small $^{2}$School of Mechanical Science and Engineering, Huazhong University of Science and Technology}\\
{\small $^{3}$Omni AI}\\[0.35em]
{\small \href{mailto:wangkailin417@gmail.com}{\texttt{wangkailin417@gmail.com}}\quad
\href{mailto:jiehaoxiang@bgyfw.com}{\texttt{jiehaoxiang@bgyfw.com}}}
}
\hypersetup{pdftitle={RegenHarness: A Robot Agent Harness with Evidence-Gated Recursive Self-Improvement},pdfauthor={Kailin Wang, Haoxiang Jie, Yaoyuan Yan, Zhiyou Heng, Zhaosong Li},pdfsubject={Robot agent harness and evidence-gated recursive self-improvement}}
\date{}

\begin{document}
% Sentence endings should not acquire wider spaces in narrow columns.
\frenchspacing
\raggedbottom
\maketitle

\begin{abstract}
Long-horizon robot execution requires a clear distinction between a model's
proposal, a controller's termination, and verified task completion. We present
\system, an evidence-gated robot-agent harness connecting task planning to
heterogeneous robot skills. Its execution architecture couples a model loop
for context-conditioned proposals with an agent loop for dispatch, observation,
verification, commitment, and bounded recovery. Four role-isolated contexts
separate planning, supervision, verification, and recovery inputs. Versioned
memory distinguishes observed facts from accepted task progress, while an
identity- and version-bound commit gate controls updates to trusted task state.
The runtime combines duplicate-dispatch control, resource leases, and recovery
budgets under explicit backend contracts, and checks the original user goal
before reporting completion. To our knowledge, we are the first to introduce
an evidence-gated recursive self-improvement (RSI) protocol for embodied
robotic agents. Across missions, execution records motivate
candidate changes to context rules, task templates, routing, and recovery
policies; fixed regression checks and release authorization govern their
acceptance; versioned rollout and rollback preserve configuration traceability.
This RSI protocol revises the harness configuration without online model-weight
updates or permission to weaken the commit gate. A real quadruped deployment
documents voice-triggered warehouse navigation, panoramic inspection, visual
analysis, message delivery, return, and spoken reporting through linked audio,
images, trajectories, and receipts. A separate circuit demonstrates why
completion depends on execution history rather than endpoint proximity alone.
Together, the cases demonstrate integrated perception, physical execution,
communication, and history-dependent completion in real-world robot tasks.
\end{abstract}


\section{Introduction}

Modern robot systems can draw on broad language reasoning, generalist
vision-language-action (VLA) policies, persistent spatial representations, learned world
models, task-and-motion planning (TAMP), and mature navigation and control stacks.  This
expanding capability pool changes the central systems question.  Rather than asking one
model to produce every motor command, a long-horizon robot agent must decide which
capability to invoke, ground that invocation in the current world, observe its progress,
determine whether its intended effect occurred, and recover when it did not.

Here a \emph{skill} is a bounded operation such as navigating to a named place or
capturing an image. A \emph{provider} (also called a backend) is the model,
controller, or service implementing that operation. A \emph{harness} is the runtime
around these providers: it decides what may be invoked, tracks execution, and
determines what counts as accepted progress. It is neither the language model
itself nor a replacement for the robot's low-level control stack.

Our starting point is HROS (Harness Robotic OS)~\cite{yan2026hros}, our previously implemented
robot operating system and inspection application. \system{} targets an upgrade
of HROS's agent harness, motivated by its deployment experience. HROS supplies
the application setting; \system{} specifies reusable execution contracts and
authority boundaries. For example, ``inspect the storage area and return''
requires navigation, observations, analysis, and reporting. Reaching the storage
waypoint proves only one of those stages, not the complete user request.

Reasoning--acting methods such as ReAct interleave model reasoning with environmental
interaction~\cite{yao2023react}.  Agent engineering also emphasizes feedback-driven tool
use and stopping conditions~\cite{anthropic2024agents}, while harness engineering makes
state, tool execution, and orchestration explicit responsibilities outside the
model~\cite{trivedy2026harness,gu2026scaling,ning2026codeharness}.  For robots, this distinction raises an additional
question: what authorizes the transition from a model's proposed progress to accepted
physical task progress?  We use \emph{model loop} and \emph{agent loop} to distinguish
inference-level proposal generation from the runtime process that acts on and verifies
those proposals, rather than to claim that iterative agent execution is itself new.

This problem becomes acute as task length grows.  For a fixed sequence of $N$ necessary
stages, let $p_i$ be the probability of stage $i$ succeeding conditional on all preceding
stages having succeeded.  The probability that all stages succeed is then
\begin{equation}
    P(\text{all stages succeed}) = \prod_{i=1}^{N} p_i.
\end{equation}
This chain-rule identity requires no independence assumption.  Recovery can change the
conditional outcomes and the execution path.
Long-horizon reliability therefore cannot come only from increasing the quality of a local
policy.  The system must also detect failed transitions, preserve useful intermediate
state, select an appropriate fallback, and prevent an early false-positive completion from
corrupting every downstream decision.

Recent systems already organize skills, memory, and verification within embodied
runtimes~\cite{zhou2026holoagent,liu2026phyagentos}. EmbodiedSkills checks skill proposals
before execution and outcomes afterward~\cite{wang2026embodiedskills}; CommitFlow gates
dependent actions on physical conditions and rechecks them after local
correction~\cite{zhao2026commitflow}. Thus, neither an execution loop nor verification
alone is unique to \system. Our focus is their runtime contract: (i) which component
may change trusted task state; (ii) which evidence may flow to the planner, skill,
verifier, and recovery policy; and (iii) how accepted progress and recovery expenditure
can be reconstructed from durable, identity- and version-bound records.

We introduce \system, an evidence-gated robot-agent harness centered on these authority and
state-transition boundaries. The key abstraction is a versioned skill invocation
bound to explicit preconditions, target
effects, resource requirements, evidence requirements, cancellation semantics, and a
recovery policy.  \system{} uses the same abstraction for a learned VLA, a navigation action,
a TAMP solver, a world-model-assisted policy, or a conventional controller.
A cross-mission recursive self-improvement (RSI) loop uses execution records
to revise the harness configuration under fixed acceptance checks, without
changing active missions or weakening their verification and safety gates.

The paper makes five system contributions:
\begin{enumerate}
    \item \textbf{Evidence-gated agent-loop control.} We distinguish model proposals,
    controller termination, and accepted task progress. Only independently verified
    outcomes may advance committed task state.

    \item \textbf{Role-isolated context compilation.} The compiler builds four bounded
    views from shared versioned memory: planner (C1), skill (C2), verifier (C3), and recovery
    (C4).  In particular, the verifier does not inherit the planner's completion claim.

    \item \textbf{Policy-neutral supervised execution and recovery.} Capability cards,
    idempotent dispatch, leases, timeouts, cancellation, and typed, budgeted recovery
    compose heterogeneous backends without granting them global-success authority.

    \item \textbf{Event-sourced auditability.} Version-bound observations, invocations,
    evidence, verdicts, commits, and recovery costs support deterministic inspection
    and restart-oriented recovery.

    \item \textbf{Evidence-gated cross-mission self-improvement.} To our knowledge,
    we are the first to introduce an evidence-gated recursive self-improvement
    (RSI) protocol for embodied robotic agents. It uses trace-driven configuration revision,
    fixed acceptance checks, mission-boundary release, and versioned rollback.
\end{enumerate}

\section{Problem Formulation}

We consider a robot receiving a long-horizon goal $g$ under partial observability.  The
robot has access to a set of heterogeneous skill providers $\Pi=\{\pi_1,\ldots,\pi_K\}$.
A provider may be a learned visuomotor policy, navigation stack, TAMP system, perception
model, remote service, or deterministic controller.  Providers differ in embodiments,
observation and action spaces, latency, risk, failure modes, and available feedback.

The harness compiles $g$ into a task contract
\begin{equation}
\mathcal{T}=(g,\mathcal{G},\mathcal{C}_g,\mathcal{S},\mathcal{R}),
\end{equation}
where $\mathcal{G}$ is an acyclic dependency graph of subgoals, $\mathcal{C}_g$ is the set of
original completion conditions, $\mathcal{S}$ contains safety and resource constraints, and
$\mathcal{R}$ defines recovery budgets.  Each node $v\in\mathcal{G}$ declares one or more
target effects $E_v$, eligible capability classes, dependencies, and evidence requirements.

\paragraph{Scope of the notation.}
A task is the complete user request; a node is one dependency-constrained stage;
an invocation is one attempt to execute a node with a particular provider.
Retries can create new invocations without changing the node's intended effect.
A node is eligible only when its required dependencies have been accepted.
The integer $k$ below is a logical state version, not elapsed time, distance,
or a model-training iteration.

At state version $k$, the system maintains two fact sets.  $O_k$ contains recent observed
facts whose provenance and freshness are known but whose task meaning may be unverified.
$C_k$ contains committed facts admitted by the verification path. A \emph{commit}
means accepting a checked result into the authoritative task record, not issuing
a motion command. For example, a pose sample belongs to $O_k$; the accepted
statement that the current navigation node reached its specified target belongs
to $C_k$. ``Committed'' describes acceptance by the configured checks, not a
guarantee that sensors or verifiers are infallible. A skill invocation may
read from an authorized context derived from $(O_k,C_k)$, but it cannot write $C_{k+1}$
directly. Instead, execution produces an evidence packet $Z_k$ and a proposed
fact update $\widehat{\Delta}_k$. The verdict vocabulary distinguishes
\emph{satisfied}, \emph{violated}, \emph{ambiguous}, \emph{stale}, or
\emph{unsafe}.
Satisfied means the declared conditions passed; violated means
a condition failed; ambiguous means the evidence cannot decide;
stale means it is no longer valid for the decision; and unsafe
denotes a safety rejection. The reference predicate verifier returns the first
three; freshness, identity, and safety checks additionally gate acceptance.
Only a satisfied verdict with matching task, node, invocation, and state-version
identities may advance committed facts. Other outcomes do not accept the proposed
facts, although recording an observation, rejection, or recovery expenditure can
still advance the overall state version.

Our objective is not to optimize a specific policy.  It is to maximize verified task
completion while minimizing false completion, unsafe actions, unrecoverable loops, resource
conflicts, and irreproducible decisions under a fixed pool of perception and action
capabilities.

\section{Design Principles and Threat Model}

The failures considered here are erroneous completion claims, stale or misbound
evidence, repeated requests, concurrent resource use, and interrupted execution.
The harness is intended to contain these runtime failures, not to prove correctness
against compromised sensors, malicious hardware, or arbitrary verifier defects.

\paragraph{Explicit authorities.}
Planning, prediction, and execution do not confer authority to verify, commit,
or override safety. A planner can propose a skill; a skill can report progress or terminal status;
a world model can predict likely effects; none of them can alone certify that a physical
goal has been achieved.  A safety veto is monotonic: later components cannot relax it.

\paragraph{Physical skills are not ordinary software tools.}
A function return usually marks the end of computation.  For a physical skill, a return can
mean that a command was accepted, a trajectory was generated, or a controller stopped.  It
does not necessarily mean that the intended object relation, robot pose, or human-facing
outcome now holds.  The harness must therefore model asynchronous progress, cancellation,
quiescence, physical evidence, and side effects.

\paragraph{Fail closed under missing evidence.}
Missing evidence, a missing verifier, stale context, or unresolved predicates must not be
coerced into success.  Ambiguity is a first-class outcome that commonly triggers active
re-perception instead of task-state advancement.

\paragraph{Bound every autonomous loop.}
Every retry, view change, policy switch, or replanning action consumes an explicit budget. This
prevents a capable planner from creating an unbounded physical loop and makes escalation to
a human or safe stop inspectable.

\section{\system{} Architecture}

\begin{figure}[!htbp]
\centering
\includegraphics[width=\textwidth]{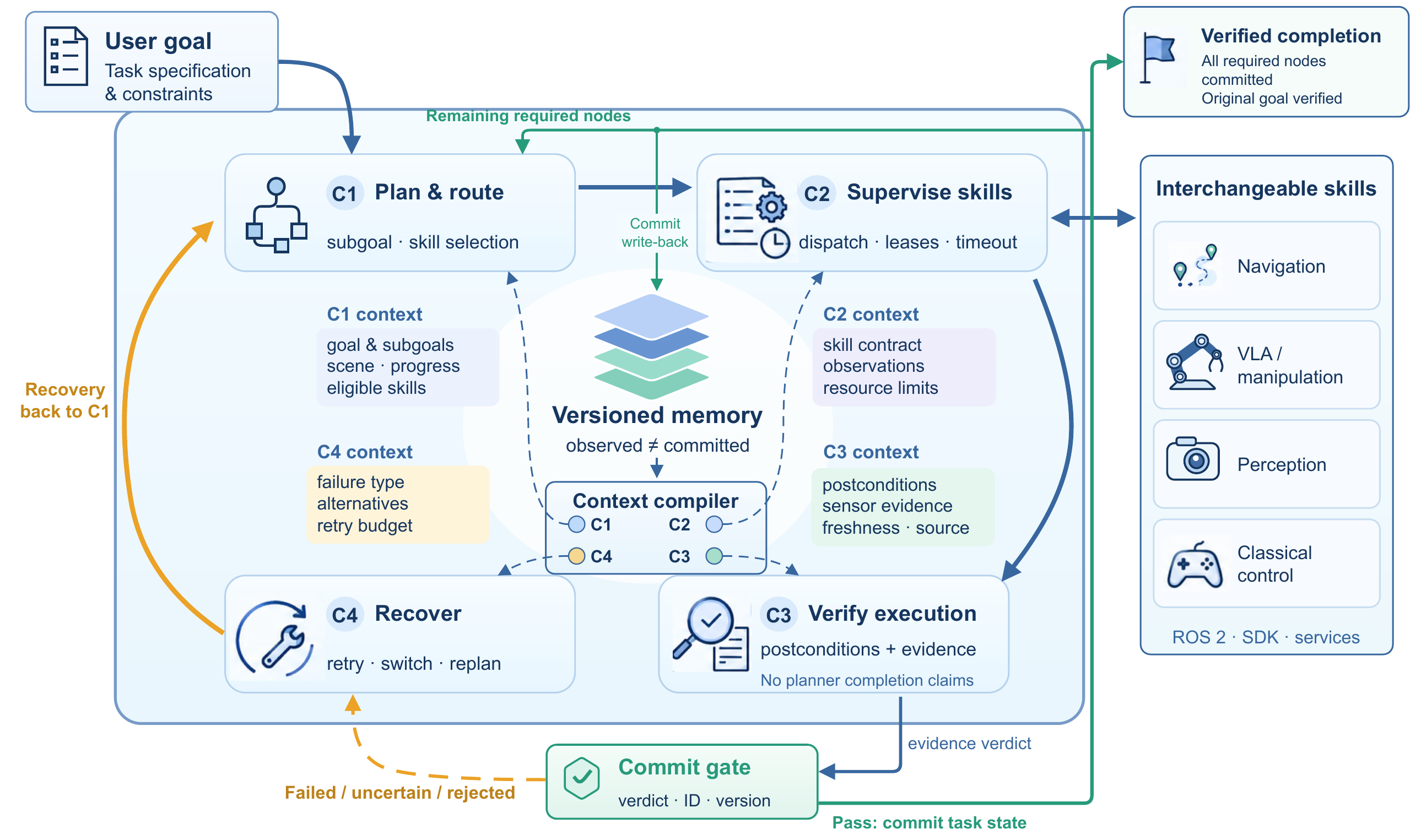}
\caption{\system{} architecture and execution authority. Versioned memory supplies
role-isolated contexts for planning (C1), supervision (C2), verification (C3), and
recovery (C4). Only the identity- and version-bound commit gate accepts task progress.
Blue paths carry execution and context; green paths carry accepted progress;
amber paths denote bounded recovery. Interchangeable providers remain outside
task-state commit authority.}
\label{fig:architecture}
\end{figure}

\subsection{Model and Agent Loops}
\label{sec:loops}

We distinguish two coupled loops by their outputs and authority, not by requiring
separate models or fixed execution frequencies.  The \emph{model loop} refers here to
inference-time proposal generation and revision as context and feedback change; it is
not a model-training loop or a world-model rollout.  The \emph{agent loop} is the
task-level process that selects, executes, verifies, and commits physical work.
The harness implements the state, interfaces, and guards connecting these loops.

\paragraph{Model loop: context to proposal.}
For runtime state $S_k$ and role $r$, a model-backed component consumes a compiled
context and emits a proposal:
\begin{equation}
c_k^{(r)}=\kappa_r(S_k),\qquad
u_k^{(r)}=M_{\theta_r}(c_k^{(r)}).
\end{equation}
Here $S_k$ includes node states, observed and committed facts, evidence references,
and remaining budgets. The role $r$ is planning, supervision, verification, or
recovery; $\kappa_r$ selects its permitted input fields, $c_k^{(r)}$ is that input,
and $M_{\theta_r}$ is a model with parameters $\theta_r$. Its output $u_k^{(r)}$
may be a subgoal, skill selection, or recovery suggestion. Not every role requires
a model: a predicate verifier is a valid deterministic implementation.
On a later invocation, refreshed observations,
verified progress, or failure evidence can change the proposal without changing model
weights $\theta_r$.  A provider may implement additional internal inference steps, but
the harness does not assume access to hidden reasoning or require a universal internal
loop.  A deterministic planner or controller can occupy the same interface.  Neither
additional reasoning nor a model's declaration of success directly updates committed
task facts.

\paragraph{Agent loop: proposal to verified progress.}
At the task level, C1 selects the next eligible node and provider, C2 supplies the
execution context, and the supervisor dispatches and polls the skill.  C3 evaluates
postconditions using execution evidence.  The commit gate admits only a valid,
identity- and version-matched result; failed or uncertain outcomes enter C4 for
budgeted recovery or escalation.  Accepted progress is written back before subsequent
contexts are compiled.  All required nodes being committed triggers an original-goal
check rather than immediate completion.  Thus, tool return, node acceptance, and
task completion are distinct stopping conditions.

\paragraph{Coupling through context and authority.}
Figure~\ref{fig:architecture} makes the coupling concrete: the central compiler delivers
bounded role views, while execution, verification, commitment, and recovery form the
outer task loop.  The loops are not synchronized one-for-one: a long-running skill may
need many controller updates and runtime polls between planning decisions.  Their
shared interface is the versioned task and evidence contract, not a transcript treated
as trusted state.  This is the scope of our embodied harness design: models may revise
proposals, but only the guarded agent loop may accept physical task progress.

\subsection{Typed Effects and Task Contracts}

An effect is a task-relevant change that can be predicted, observed, and tested.  It may be
symbolic (\texttt{held(cup)}), metric (the base lies within a pose tolerance), relational
(\texttt{inside(cup,drawer)}), visual (a target mask or keypoint configuration), or temporal
(a condition remains stable for a required interval).  Multiple representations can be
bundled when no single representation is sufficient.

A task contract retains the original user goal and completion conditions while organizing
execution into a dependency graph.  Each node contains a subgoal, success conditions,
eligible capability classes, typed parameters, resource requirements, evidence policy,
timeout, and recovery limits.  This separation is important: the graph records what must
become true, while routing determines which currently available backend should attempt it.

``Typed'' means that fields have explicit categories and validation rules rather
than being an unconstrained prose instruction. A navigation target, for example,
must identify a coordinate frame, a target position, and an admissible tolerance.
Its preconditions apply before dispatch (such as a valid localization estimate);
its postconditions are tested afterward (such as reaching the target). A request
to inspect and return additionally requires evidence from the observation and
reporting nodes. These task conditions are not interchangeable with a controller's
terminal status.

\subsection{Capability Cards and Routing}

Each skill provider publishes a capability card containing its identifier and version,
supported embodiments, parameter schema, preconditions, declared side effects, expected
latency, resource set, cancellation behavior, risk class, failure vocabulary, and required
evidence.  Routing begins with deterministic filtering over capability, embodiment,
parameter validity, preconditions, freshness, and safety.  A configurable scorer may then
rank the eligible set using predicted success, risk, uncertainty, latency, cost, and verified
operational history:
\begin{equation}
s(\pi_i)=w_p\hat p_i-w_r\hat r_i-w_u\hat u_i-w_l\hat l_i-w_c\hat c_i
            -w_h\hat h_i.
\end{equation}
Here $\hat p_i,\hat r_i,\hat u_i,\hat l_i,\hat c_i$ denote estimated success,
risk, uncertainty, latency, and cost for provider $i$; $\hat h_i$ is a penalty
derived from relevant historical failures or transition residuals. A transition
residual is a discrepancy between an expected and observed effect. Nonnegative
weights $w_*$ specify application tradeoffs after the quantities are put on
compatible scales. This is a configurable ranking interface, not a learned
policy or a scoring rule calibrated by the navigation cases in this paper.
The score is inspectable and is not itself an execution authority; hard guards and resource
acquisition still run after ranking.

This interface makes broad VLA policies and precise classical stacks complementary.  A VLA
can supply a flexible manipulation backend, while a navigation stack or TAMP solver handles
tasks where metric constraints and explicit collision checks dominate.
Provider fallback preserves the node's target effect.

\paragraph{Optional world-model support.}
A world model can estimate candidate outcomes to assist routing or local planning.
Its predicted effects remain decision inputs: they cannot establish post-execution
success.  The harness can also operate with geometry-based planners, navigation skills,
or conventional controllers that supply no learned prediction.  Thus, the task contract,
context compiler, execution supervisor, and commit path do not require a particular
world model.  This paper studies those runtime mechanisms; prediction-model training
is delegated to the selected provider.

\subsection{Role-Isolated Contexts}

\system{} compiles a common state version into four bounded contexts (Table~\ref{tab:contexts}).
The letter C stands for \emph{context}; C1--C4 are information-access roles,
not four model names, robots, or independently maintained memories. A context
compiler is a software selector and formatter for those views, not a neural
network compiler. ``Bounded'' refers to the scope and amount of information
included in a view.
The compiler selects fields by role, task and node scope, confidence, freshness, provenance,
and item budget.  It rejects expired or superseded records and withholds unresolved
conflicts until re-observation or explicit disambiguation.

\begin{table}[t]
\centering
\small
\begin{tabular}{@{}lll@{}}
\toprule
View & Consumer & Primary question \\
\midrule
C1 & Planner & What should happen next? \\
C2 & Skill & How should this node be executed? \\
C3 & Verifier & Did the intended effect occur? \\
C4 & Recovery & What bounded response is justified? \\
\bottomrule
\end{tabular}
\caption{Role-isolated contexts compiled from one versioned state.}
\label{tab:contexts}
\end{table}

C1 contains the original goal, graph frontier, committed progress, relevant spatial state,
and high-level failure history.  C2 contains one current subgoal, grounded targets, the
selected skill contract, bounded observations, and a small number of verified operational
examples. C3 contains postconditions, before/after evidence, sensor timestamps, and
relevant committed facts; records originating from the planner's own conclusion are
excluded.  C4 contains the structured failure, attempted actions, remaining budgets,
available alternatives, and evidence necessary to choose an escalation.

Context separation limits both accidental leakage and correlated self-confirmation.  For
example, the text ``the cup has been grasped'' generated by a planner is useful as an
intention in C2 but cannot become evidence of grasp success in C3.
In this sense, context engineering is an information-access policy for the coupled loops,
not merely prompt formatting: each consumer receives the state and evidence permitted
for its decision, rather than an undifferentiated history of model outputs.

\subsection{Supervised Skill Lifecycle}

The skill supervisor is the only dispatch path in the reference harness. Before execution, it validates the request,
records a dispatch intent, acquires leases for the robot and workspace resources, and binds
an idempotency key to the task/node/attempt.  The backend then exposes a lifecycle with
accepted, queued, running, blocked, succeeded, failed, and cancelled states, plus structured
progress and evidence references.

\emph{Idempotent dispatch} means that receiving the same invocation request
again returns its existing execution record rather than starting another
physical action. This handles a repeated send after a lost acknowledgment; it
does not prohibit an explicitly authorized new attempt. The reference
orchestrator keys dispatch by task, node, and state version. A \emph{resource
lease} is a time-bounded reservation, with an owner and expiry, for resources
such as the mobile base. Conflicting exclusive leases prevent two skills from
simultaneously commanding that resource. A lease is coordination metadata, not
a physical emergency stop.

Long-running skills have explicit deadlines.  Persisted execution handles support
reattachment to providers that implement it.  Replaying a pending start uses the same
request identity and requires provider-side deduplication to avoid repeated physical
execution across that crash boundary.

For physical deployment, resource reassignment requires a confirmed stopped or safe-hold
state.  The reference supervisor currently treats return from \texttt{cancel()} as
completion and releases the lease.  This assumes synchronous cancellation; asynchronous
robot providers require a stopping acknowledgment before the lease can be reassigned.

\subsection{Evidence, Verification, and Commit}

For a terminal execution, the harness preserves an evidence packet, summarized as:
\begin{equation}
Z_k=\bigl(I,k_{\mathrm{ctx}},O_{\mathrm{before}},O_{\mathrm{after}},P,\widehat{\Delta}_k\bigr),
\end{equation}
where $I$ identifies the task, node, provider, execution, and packet;
$k_{\mathrm{ctx}}$ is the state version used to dispatch; $O_{\mathrm{before}}$
and $O_{\mathrm{after}}$ reference observations before and after execution;
$P$ references pose or other state estimates; and $\widehat{\Delta}_k$ contains
the provider's proposed task facts. An observation collection may be empty if
that skill does not require it; its evidence policy determines what is mandatory.
Each evidence reference retains a source, timestamp, confidence, and a file
reference or checksum, with a coordinate frame where relevant. In the reference
implementation, the packet ID is derived from the execution ID, which is also
stored in packet metadata. The packet is retained even when verification rejects it.

Verification proceeds in two layers.  A node-level verifier evaluates every declared
postcondition with sources independent of the executing provider where possible. A
backend's own \texttt{done} signal may be included as evidence but is never the sole
authority.  The commit gate checks the verdict, identity binding, state version, and
condition coverage before atomically advancing the node and trusted facts.  Once every
required node is committed, a separate goal verifier rechecks the original completion
conditions.  Thus:
\begin{equation}
\text{controller done}\not\Rightarrow\text{node success}
\not\Rightarrow\text{goal success}.
\end{equation}
Only a satisfied original-goal verdict authorizes a final success report to the
user; intermediate status or failure reports are not completion claims.

The verifier asks whether the evidence satisfies the node conditions; the commit
gate asks whether that verdict is authorized to change this task version.
For example, a valid arrival verdict from an earlier attempt cannot certify a
later request to a different target. The gate rejects a packet generated from
an obsolete state version rather than silently interpreting it under the new
state. ``Atomic'' means that accepting the node result and updating its task
facts form one guarded state transition, not two independently visible writes.

\subsection{Budgeted Recovery}

Non-satisfied verification results enter C4 with a structured failure class.  The recovery
router maps this class to an ordered escalation ladder, for example:
\begin{align*}
\texttt{target\_not\_found} &:\ \text{re-perceive}\rightarrow\text{adjust view},\\
\texttt{no\_progress} &:\ \text{local replan}\rightarrow\text{switch skill},\\
\texttt{unsafe\_action} &:\ \text{safe stop}\rightarrow\text{human handoff}.
\end{align*}
Each action consumes both a per-action and total recovery budget.  The current failure,
prior attempts, excluded providers, evidence, and remaining budget are recorded before the
next attempt is planned.  Repeating the same failed action without new evidence can
therefore be prohibited deterministically.

Recovery budgets count authorized recovery actions, not low-level controller
cycles. A retry repeats a bounded skill invocation, a provider switch chooses
another implementation of the same capability, and replanning revises the task
or route proposal. The relevant limits come from the task contract; exhausting
them requires escalation or termination instead of another automatic attempt.
An internal collision-avoidance iteration is therefore not, by itself, a C4 event.

\subsection{Trusted Memory and Event Sourcing}

Memory here means stored task and evidence records, not a learned model of how
the world will evolve. Versioning marks which record was current for a decision;
it does not change model weights.

The memory fabric separates five roles: working memory for the active graph and control
state; spatial memory for maps, entities, poses, and relations; episodic memory for verified
attempt histories; operational memory for context-conditioned provider outcomes; and
semantic memory for capability contracts and rules.  Long-term records require provenance,
confidence, validity intervals, and verification state.  Updates append a new record with a
\texttt{supersedes} or conflict relation instead of silently overwriting history.

The trusted task store projects current progress from ordered observation, commit,
recovery-budget, and goal-verification events.  Execution snapshots and evidence records
retain the associated request, provider handle, state version, sources, and verdicts.
These records support reconstruction of task progress.  Exact replay of model decisions
additionally requires the original context contents, candidate set, configuration,
model outputs, and any randomness; complete capture of those inputs across all entry
points remains an integration requirement.

Event sourcing means retaining an ordered history of state-changing records,
from which current task progress can be reconstructed. Reconstruction is not
the same as re-running the robot or reproducing a stochastic model response.

\subsection{Evidence-Gated Recursive Self-Improvement}
\label{sec:rsi}

Recursive self-improvement (RSI) adds a cross-mission revision loop around the
model and agent loops. The model loop revises a proposal within the current
configuration; the agent loop verifies physical progress; the RSI loop evaluates
a change to the configuration used by subsequent missions. Here the object of
improvement is the harness configuration: context-selection rules, task
templates, routing preferences, and bounded recovery policies. The RSI protocol
organizes candidate generation, regression checks, release authorization,
and versioned rollback.

\paragraph{From execution records to candidate revisions.}
Let $H_j$ denote harness configuration version $j$, and let $\mathcal{D}_j$
contain its mission records, evidence links, failures, and recovery costs.
A revision generator produces a candidate $\widetilde H_j$ with a change
manifest that identifies the affected component, motivating traces, intended
effect, and rollback version. Stable skill and evidence interfaces allow the
same provider to be exercised under both configurations.

\paragraph{Evaluation before release.}
Each candidate is checked against a fixed regression set $\mathcal{R}_j$,
including nominal tasks and stale-evidence, duplicate-request, interrupted-run,
and resource-conflict cases. The release rule is
\begin{equation}
H_{j+1}=\begin{cases}
\widetilde H_j, & \mathcal{A}(\widetilde H_j,H_j,\mathcal{R}_j)=1,\\
H_j, & \text{otherwise}.
\end{cases}
\end{equation}
The acceptance predicate $\mathcal{A}$ requires preserved safety and evidence
contracts, no regression on designated checks, improvement on a predeclared
target metric, and release authorization. Evaluation results and tested versions
are retained with the revision. The updater has no permission to weaken its own
acceptance predicate, commit gate, or lower-level safety interlocks.

\paragraph{Versioned rollout and recursion.}
Approved revisions become available at mission boundaries; each active mission
remains pinned to its original configuration. A failed release returns to the
last accepted version. Subsequent mission records feed the next revision cycle,
forming the recursive feedback path. Online weight updates are outside this
protocol. Section~\ref{sec:extension-evaluation} defines the assessment
criteria for execution outcomes and cross-mission configuration updates.

\subsection{Safety and Resource Authority}

The harness does not replace motor-control safety.  Hardware limits, collision and force
guards, watchdogs, controller interlocks, and emergency stops remain beneath and outside
model authority.  The harness adds task-level preflight, resource and workspace leases,
freshness checks, cancellation, and human-approval policies.  A planner or plugin may add a
restriction but cannot remove a restriction imposed by a lower safety layer.  Missing
safety-critical state causes the relevant action to fail closed.

\section{Implementation}

The Python reference runtime implements typed immutable contracts and
SQLite-backed task, evidence, execution, and memory stores. Its state machine
performs planning, context compilation, dispatch, polling, verification,
commitment, recovery, and original-goal verification. Persisted requests,
execution handles, leases, and evidence support restart-oriented recovery under
the deduplication and cancellation assumptions stated above. Adapters connect
ROS~2 (Robot Operating System 2), vendor software development kits (SDKs),
model services, and simulators to that runtime.

The reference deployment is single-host. Multi-robot deployment requires
transactional external leases and explicit clock and event-ordering contracts.
Verification also depends on evidence quality: shared sensor or model errors
require independent adjudication when measuring semantic correctness.

Section~\ref{sec:evaluation} presents the quadruped warehouse workflow built
on the HROS field application~\cite{yan2026hros}, together with exploration,
mapping, and navigation results. The provider interface also accommodates
VLA, manipulation, TAMP, and world-model-assisted backends through the
capability and lifecycle contracts defined above.

\section{Experimental Evaluation}\label{sec:evaluation}

\subsection{Experimental Setup and Robot Platform}

The main case documents a physical quadruped's voice-triggered warehouse
inspection. Voice-session records, mission identifiers,
image checksums, and navigation receipts link the stages in
Figure~\ref{fig:recorded-inspection} to one execution. A separate circuit
examines completion when the robot starts near its eventual
goal. These deployment cases demonstrate an integrated physical inspection
workflow and a history-dependent navigation completion condition.

The experiments use a quadruped robot in an office and entrance environment.
We examine the voice-to-action workflow, VLM-guided exploration with
\texttt{omni\_planner}, semantic place binding, and navigation completion.
Mission identifiers connect the command, sensor observations, action results,
and communication receipts. Section~\ref{sec:extension-evaluation} defines
the assessment criteria for harness execution and RSI configuration updates.

\paragraph{Platform and recorded tasks.}
The robot's navigation stack combines a fixed reference route, local planning, localization,
and a supervised velocity interface. The recorded trial settings specify a
0.5\,m/s speed cap and a 0.6\,m controller lookahead, the forward reference
distance used by the local follower. The first mission follows
the same 17.62\,m reference in opposite directions between the violet and blue
waypoints. The second follows an 18.83\,m circuit whose start and goal coincide.

The ROS~2 recordings use the MCAP container format. They store body poses on
\path{/vbot_planner/body_pose} and dispatched paths on \path{/initial_path}.
A reference path is the ordered waypoint sequence
sent to the follower; the body trajectory is the robot's recorded localization
output while following it. We also inspect supervisor and planner logs. Figures
use the recorded \texttt{map} frame; their background is a saved point-cloud map
aligned by the recorded static transform. Camera panels are decoded from the
closed-route recording, with timestamps derived from MCAP log time. In the
distance plots, zero seconds is the reference-path dispatch recorded in that
segment's bag. Audio acquisition and recording start have separate timestamps.

\subsection{Voice-Triggered Warehouse Inspection}

The spoken instruction translates as ``Help me go and check the office
storage area,'' preceded by the wake phrase. The warehouse profile connects
the command to navigation, panoramic capture, visual analysis, DingTalk
delivery, return, and spoken reporting. The profile supplies return and reporting
as part of its inspection workflow. Figure~\ref{fig:recorded-inspection}
presents the seven-stage execution of the voice-triggered warehouse inspection
described in this subsection.

\begin{paperanchoredfigure}
  \includegraphics[width=\textwidth,trim=0 8bp 0 45bp,clip]{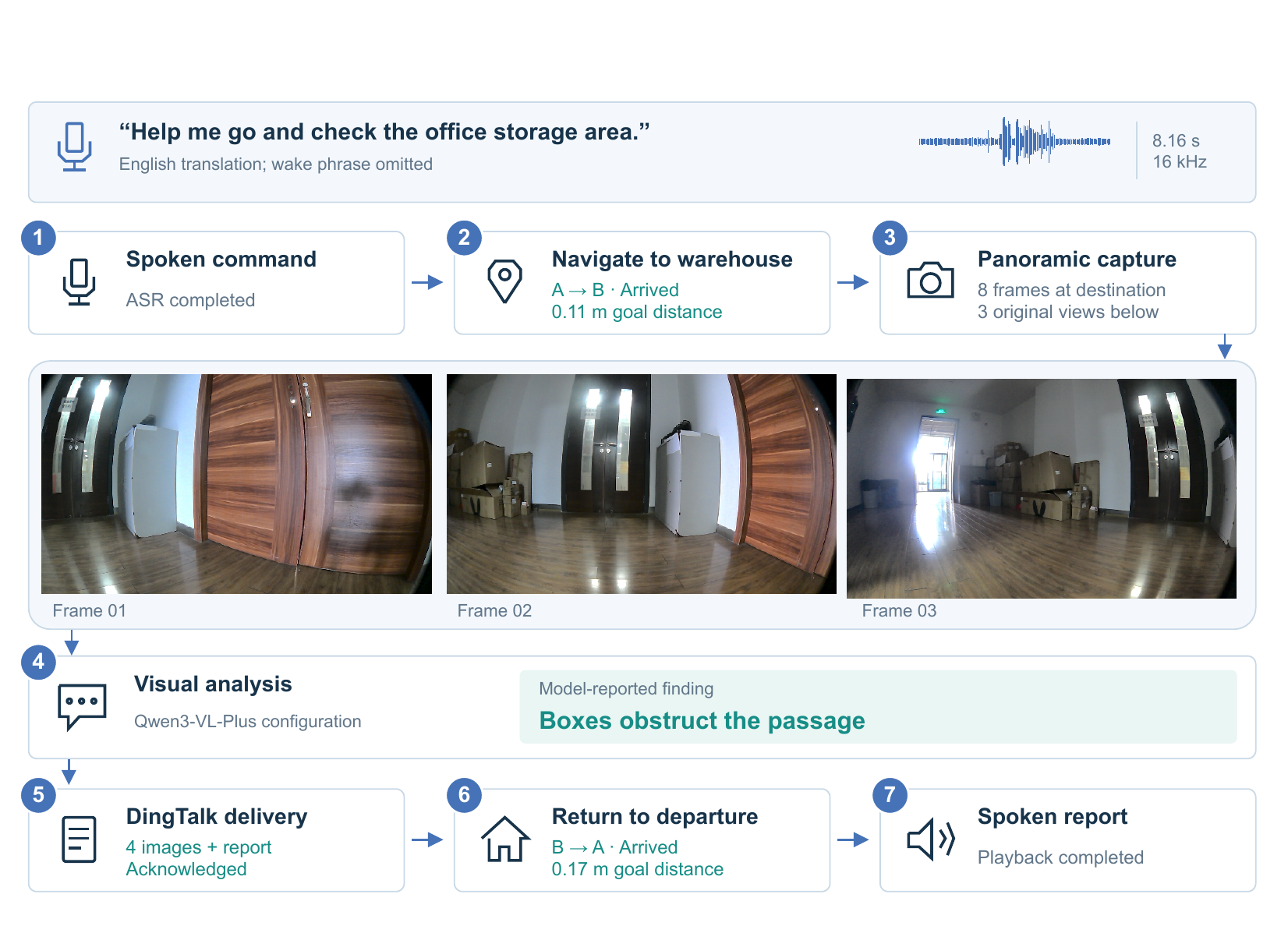}
  \caption{Step-by-step evidence from one real voice-triggered warehouse
  inspection. The waveform is derived from the original microphone recording;
  the photographs are unaltered full frames from the destination panorama.
  English text translates or summarizes the saved command and report.
  Arrival distances are supervisor-reported; communication outcomes are
  system-side receipts. Each panel summarizes the corresponding recorded stage.}
  \label{fig:recorded-inspection}
\end{paperanchoredfigure}

\paragraph{Command and navigation.}
An 8.16\,s, 16\,kHz microphone recording feeds streaming ASR. The warehouse
profile binds the destination to \texttt{office\_storage} at endpoint B.
The robot navigates from A to B; the supervisor reports 0.11\,m endpoint
error and disarms velocity control before the observation stage begins.

\paragraph{Observation and visual analysis.}
The robot rotates in segments and saves eight 1280$\times$720 images,
with capture times, checksums, rotation request identifiers, terminal events,
and a static-body postcondition. The deployed Qwen3-VL-Plus configuration
analyzes these frames. Its saved report describes boxes obstructing a
passage, providing the visual finding used by the subsequent reporting stages.

\paragraph{Delivery, return, and user feedback.}
The application sends four images and the inspection text to the configured
DingTalk group, retaining acknowledged receipts before returning to A.
The return supervisor reports 0.17\,m endpoint error. After a turn and a
static-body check, the robot speaks the result. Final playback is recorded
as completed. Delivery and playback records preserve both system-side outcomes.

\paragraph{Linking the stages.}
The voice result embeds the same task identifier and mission record as the
executor; the transcript checksum matches, navigation receipts locate both
legs' recordings, and all eight image checksums match. The session log
links the recognized command to final playback. Mission execution takes
259.94\,s; the full voice turn, including input acquisition and overhead,
takes 271.22\,s. The shared identifiers, timestamps, and checksums connect
these stages into one physical mission under the deployed warehouse profile.

\paragraph{End-to-end mission results.}

Figure~\ref{fig:recorded-inspection} follows one spoken request through
arrival, observation, analysis, delivery, return, and reporting. The
photographs show the actual destination and boxes described in the output.
The images retain the camera's original lighting and lens distortion.
Navigation, image capture, messaging, and speech each supply
different evidence. Their ordered outcomes make the complete workflow
visible: reaching the warehouse alone would leave inspection, delivery,
return, and final reporting unfinished.

\paragraph{Cross-stage evidence links.}

The voice result embeds the mission identifier and the executor's record.
The transcript checksum matches the input; navigation receipts locate both
legs' recordings; frame paths match saved image checksums. The final record
retains the finding and delivery status. These links connect audio, motion,
perception, and communication in one physical task. The case documents the
application's complete execution chain and identifies the evidence produced
at each stage. Shared identifiers and evidence references preserve the
connection between the user request, physical actions, visual findings,
message delivery, and final spoken report.

The warehouse mission links the following evidence across its execution stages:
\begin{itemize}
  \setlength{\itemsep}{2pt}\setlength{\parskip}{0pt}
  \item \textbf{Input:} the original recording, ASR transcript, and a
  matching transcript checksum in the mission record.
  \item \textbf{Inspection:} timestamped panoramic images, rotation
  receipts, and a visual report based on those frames.
  \item \textbf{Physical execution:} outbound and return trajectories,
  arrival declarations, and recorded static-body checks.
  \item \textbf{Communication:} acknowledged image/text delivery and
  completed final speech playback.
\end{itemize}

\subsection{Exploration with VLM + \texttt{omni\_planner}}

Mapping supplies spatial context for warehouse navigation and semantic place
binding. Figure~\ref{fig:exploration-mapping} presents the VLM +
\texttt{omni\_planner} workflow alongside the field map. The exploration
system uses the VLM for semantic goal selection and our in-house
\texttt{omni\_planner} for local motion generation. SLAM maintains spatial
context, and execution feedback informs the next exploration decision.

\begin{paperanchoredfigure}
  \includegraphics[width=\textwidth,trim=0 15bp 0 35bp,clip]{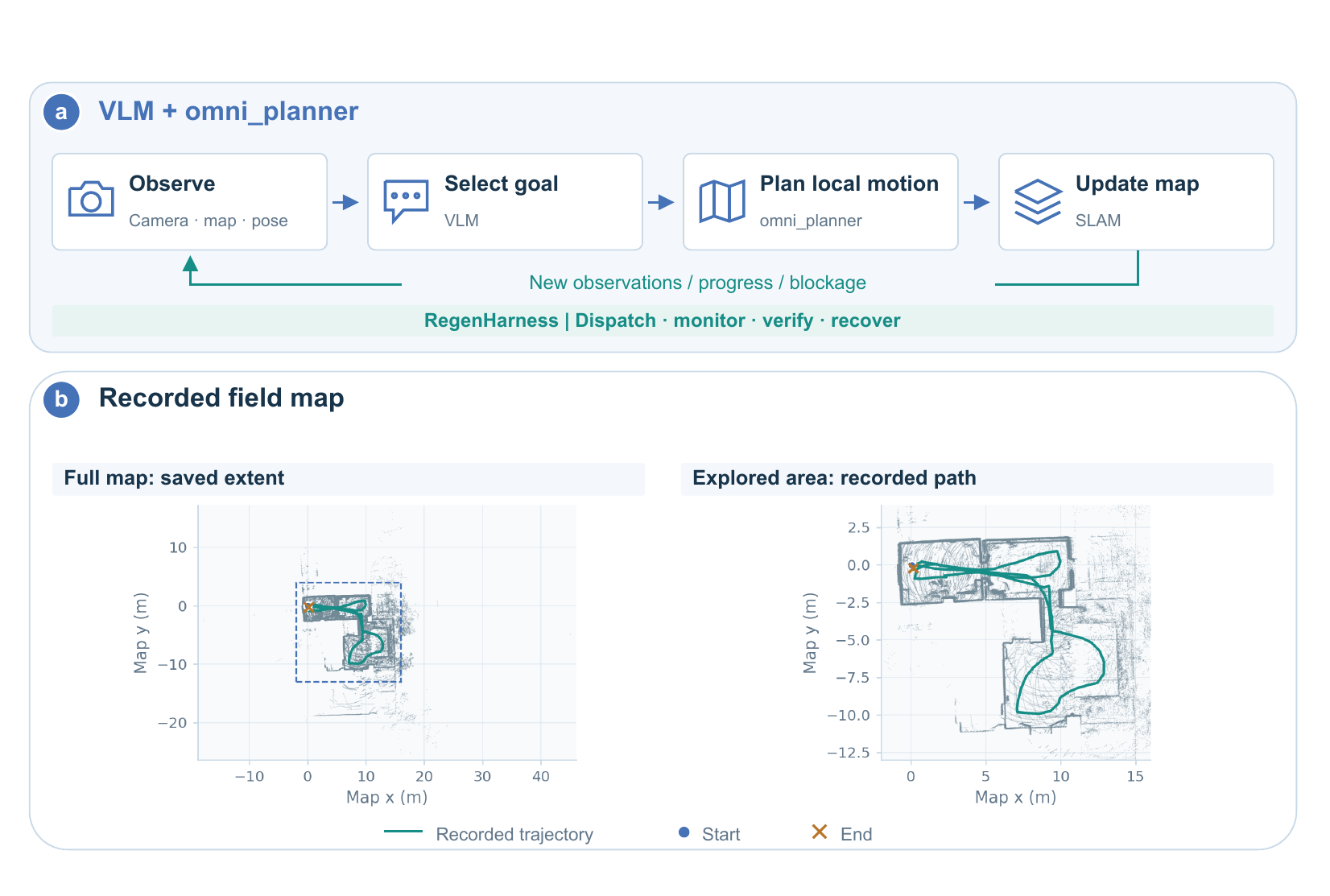}
  \caption{VLM + \texttt{omni\_planner}: exploration architecture and field
  mapping output. (a) The architecture assigns exploration-goal selection to
  the VLM, local motion generation to \texttt{omni\_planner}, and map updates
  to SLAM, with RegenHarness supervising execution. (b) The saved
  point cloud and recorded SLAM trajectory, aligned using recorded transforms.
  The two panels connect the exploration workflow to the site's mapped
  geometry and robot trajectory.}
  \label{fig:exploration-mapping}
\end{paperanchoredfigure}

\paragraph{Visual goal selection.}
The VLM interface consumes camera observations, map context, the exploration
objective, and execution feedback. Its output is a bounded exploration target
and the reason for visiting it. This target specifies the next observation
region or reachable approach point. The harness checks target admissibility
and binds the dispatched request to the active mission and configuration.

\paragraph{Local planner.}
The local planner, \texttt{omni\_planner}, converts the selected target into
locally feasible motion using geometric observations and robot pose. Local
collision checks and controller limits govern motion execution. This division
assigns semantic target selection to the VLM while retaining geometric
feasibility and motor-command authority in the local motion stack.

\paragraph{Feedback.}
Target progress, blockage, and map changes return to the goal-selection
interface. RegenHarness supplies dispatch, monitoring, verification, and bounded
recovery around each invocation. SLAM updates the spatial map during motion.
This interaction closes the feedback path between observation,
target selection, local planning, and the next observation.

\paragraph{Mapping result.}
The field panel uses the saved point cloud from the robot's exploration and
mapping session. We render the final nonempty recorded
\path{/path} trajectory in the \texttt{map} frame and apply the recorded
transform from the point cloud's \texttt{lidar\_init} frame. The full extent
and enlarged explored area expose the geometry used for subsequent navigation
and place binding. The SLAM path records the robot's motion, including revisits.

\subsection{Site Map and Semantic Anchors}

Figure~\ref{fig:semantic-site} restores the spatial context omitted by the
trajectory crops. It displays the full extent of the saved site's vertical
point-cloud structure and an enlarged task area. The field inspection template
binds \texttt{office\_storage} to the blue route endpoint B; the violet endpoint
A is the departure and return anchor. These configured task semantics bind
the inspection destination and return point to the recorded geometric map.
The letters A and B identify task roles consistently across the mission,
map, and navigation figures.

\begin{paperanchoredfigure}
  \centering
  \includegraphics[width=\textwidth,trim=23bp 33bp 24bp 62bp,clip]{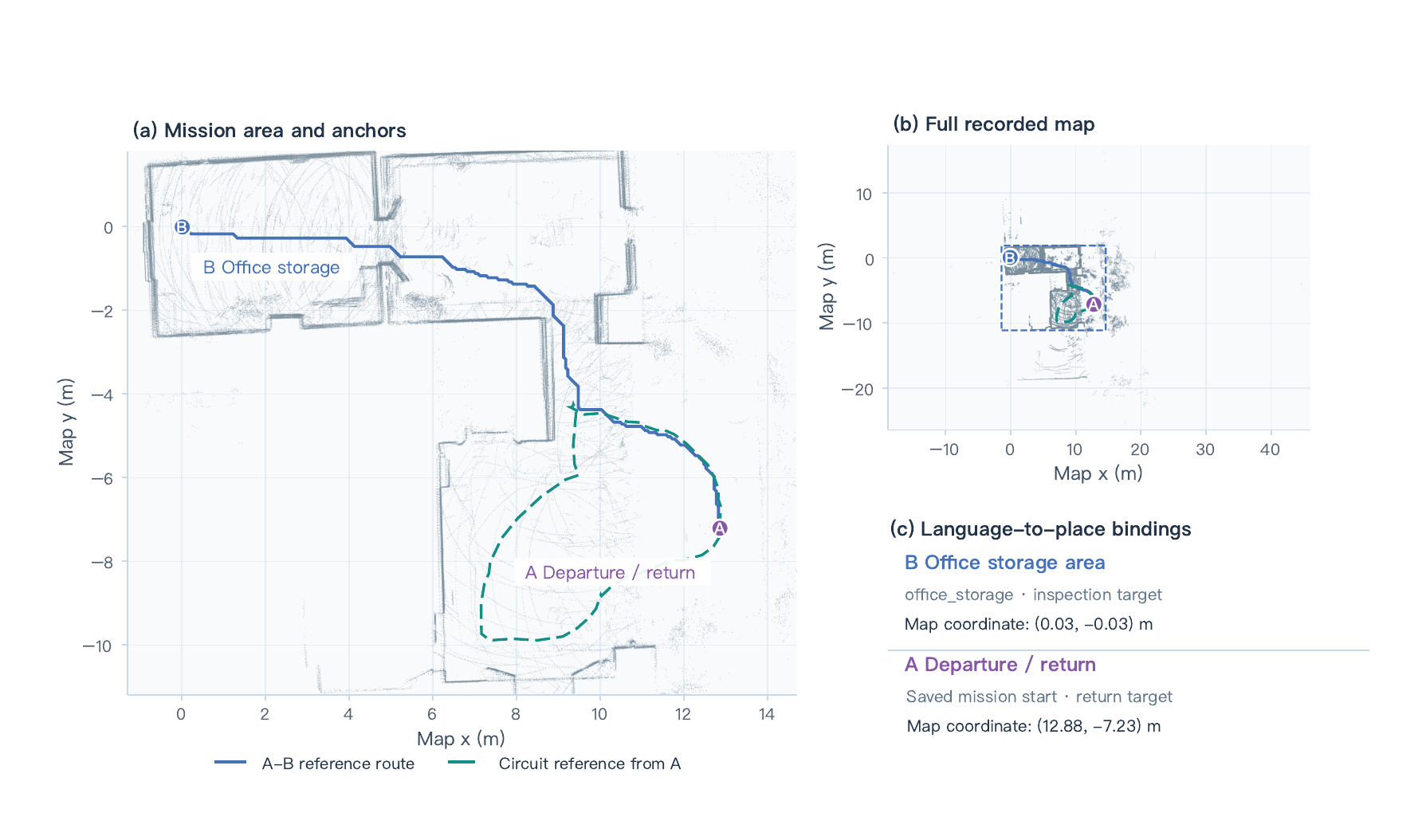}
  \caption{Site geometry and configured task semantics. (a) Task area and
  reference routes. (b) Full recorded map extent. (c) Place-to-target bindings:
  A is departure/return; B is \texttt{office\_storage}. These sparse anchors
  provide configured mission-level semantics over the recorded geometric map.}
  \label{fig:semantic-site}
\end{paperanchoredfigure}

The semantic layer connects user-level names to execution targets. Its place
record schema includes an identifier, name and aliases, map frame and version,
region or approach pose, provenance, and verification status. The field map
instantiates this connection with two configured mission anchors. Extending
the registry adds further place bindings without changing the execution contract.

\subsection{Navigation Results}

\paragraph{Metrics.}

For reference waypoints $q_0,\ldots,q_M$ and a recorded planar body position
$p(\tau)$, the quantities in Table~\ref{tab:real-cases} are
\begin{align*}
L_{\mathrm{ref}}&=\sum_{j=1}^{M}\lVert q_j-q_{j-1}\rVert_2,\\
d(\tau)&=\lVert p(\tau)-q_M\rVert_2.
\end{align*}
Here $\tau$ is elapsed time relative to recorded path dispatch, $q_M$ is the
segment's target, and $\lVert\cdot\rVert_2$ is planar Euclidean distance.

For example, the outbound row assigns a 17.62\,m A-to-B route. At the arrival
declaration the supervisor reports 0.11\,m remaining to B; the last recorded
pose is 0.09\,m from B. The two columns report the supervisor's arrival event
and the final recorded localization sample, respectively. Their distinct
sampling times explain the difference. Recording duration includes
pre/post-arrival samples; mission timing is taken from the execution log.

\paragraph{Execution results.}

Both round-trip segments reach their endpoints and then stop publishing velocity
commands (control disarming). The return trace has an interval of limited
goal-distance reduction; planner logs record rejected collision candidates and
replanning before progress resumes. These events describe the local planner's
internal motion-generation process.

\begin{paperwidefigure}
  \centering
  \includegraphics[width=\textwidth,trim=42bp 32bp 26bp 50bp,clip]{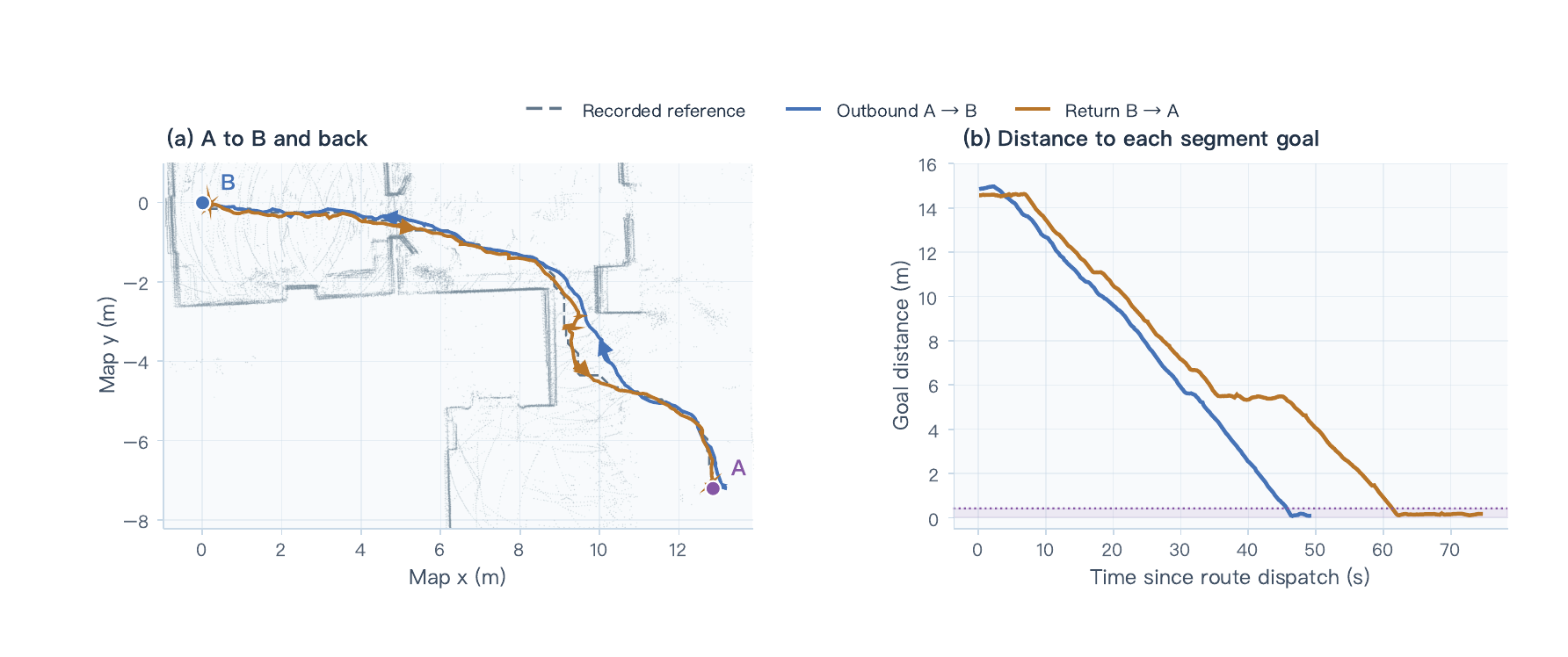}
  \caption{Navigation legs of the recorded voice-triggered warehouse mission.
  (a) Reference and body trajectories from
  A to B and back. (b) Distance to each segment's goal relative to reference
  dispatch; the dotted line is the 0.40\,m tolerance. Traces continue to the
  end of recording; Table~\ref{tab:real-cases} reports the supervisor's
  arrival distances and the distances at the last recorded poses.}
  \label{fig:real-roundtrip}
\end{paperwidefigure}

\vspace{5pt}
\begin{paperwidetable}
  \caption{Per-segment navigation measurements. Outbound and return are the two legs of the
  voice-triggered warehouse inspection; the circuit is a separate mission.
  A is the departure/return anchor and B the configured storage-area target
  in Figure~\ref{fig:semantic-site}. All numeric distances are in meters.}
  \label{tab:real-cases}
  \footnotesize
  \setlength{\tabcolsep}{7pt}
  \renewcommand{\arraystretch}{1.2}
  \begin{tabular}{@{}llrrr@{}}
    \toprule
    Execution segment &
    \shortstack[l]{Commanded route\\and target} &
    \shortstack[r]{Reference path\\length $L_{\mathrm{ref}}$ (m)} &
    \shortstack[r]{Goal distance at\\arrival declaration (m)} &
    \shortstack[r]{Goal distance at\\last recorded pose (m)} \\
    \midrule
    Outbound & A $\rightarrow$ B; target B & 17.62 & 0.11 & 0.09 \\
    Return & B $\rightarrow$ A; target A & 17.62 & 0.17 & 0.17 \\
    Circuit & A $\rightarrow$ circuit $\rightarrow$ A & 18.83 & 0.25 & 0.15 \\
    \bottomrule
  \end{tabular}
  \par\vspace{4pt}
  \begin{minipage}{\textwidth}
    \small
    \textit{How the columns are obtained.} Reference length is the sum of
    planar distances between commanded waypoints, not traveled distance.
    Arrival distance is the value printed in the supervisor's
    \texttt{GOAL REACHED} log; last-pose distance is recomputed from the final
    recorded localization sample to that segment's target. Both are endpoint
    distances, not path-tracking error or external ground-truth accuracy.
    The configured arrival radius is 0.40\,m; the circuit additionally requires
    confirmed departure before return can count. The two timestamps differ.
  \end{minipage}
\end{paperwidetable}

\subsection{Closed-Route Completion Case}

The circuit exposes a useful completion ambiguity. At reference dispatch, the
first recorded robot position is approximately 0.23\,m from the goal, already
inside the 0.40\,m arrival tolerance. Nevertheless, the intended task is to
traverse the circuit and return. The recorded supervisor announces departure
before its eventual arrival decision; the body trajectory moves as far as
6.20\,m from the shared endpoint and returns
(Figure~\ref{fig:real-loop}). The planner log additionally reports 99.6\%
reference progress and 0.068\,m remaining at its completion event. These are
complementary observations of endpoint distance and route progress at their
respective recorded times.

The circuit's completion condition requires departure, route progress, and
return, rather than endpoint proximity alone. The deployed supervisor records departure and return;
the planner reports reference progress; and the trajectory records the
traversed geometry. Together, these records explain the history-dependent
navigation decision and give the agent loop concrete evidence fields for
checking a closed-route task.

\begin{paperanchoredfigure}
  \includegraphics[width=\textwidth,trim=37bp 17bp 19bp 59bp,clip]{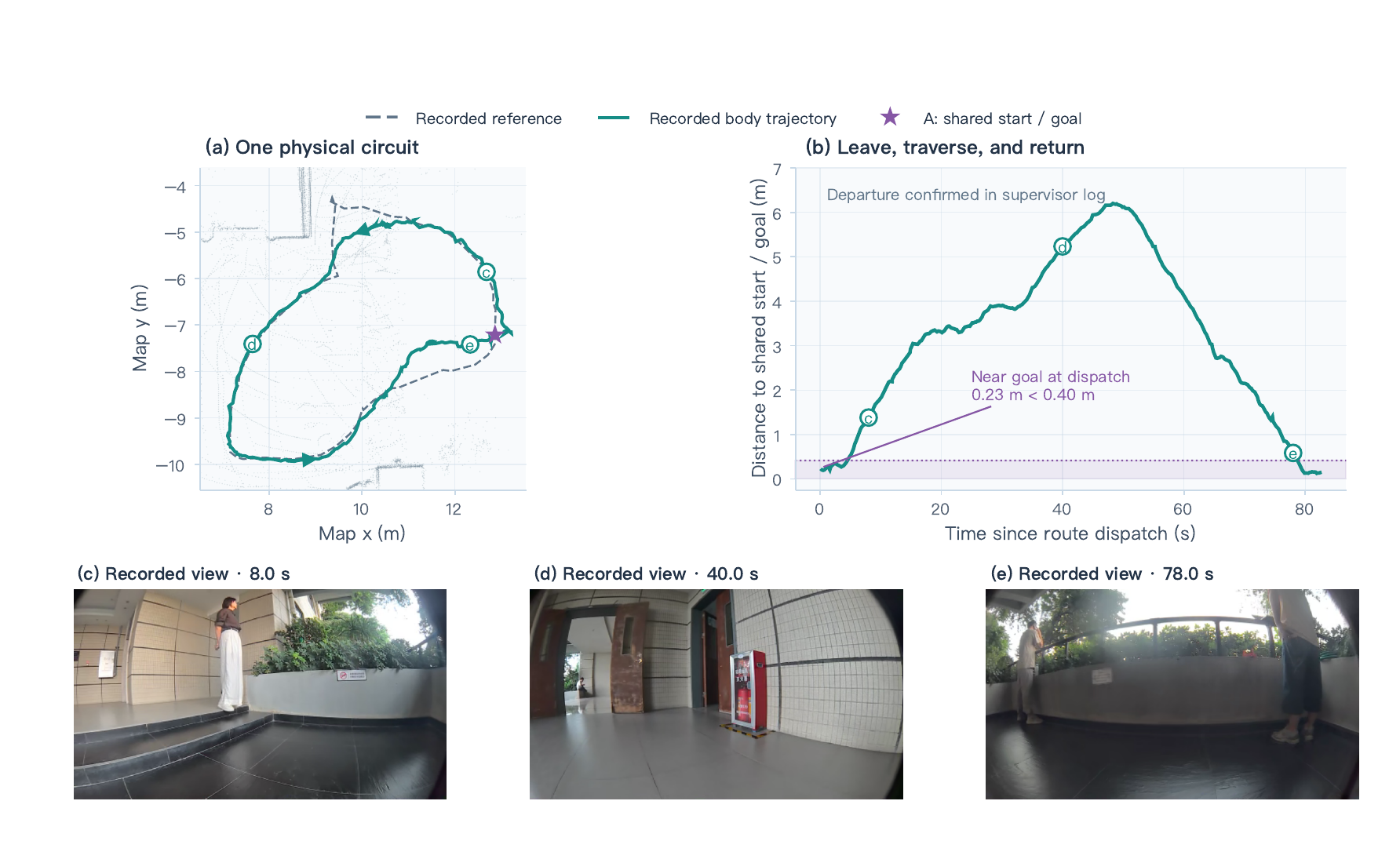}
  \caption{Circuit completion requires execution history. (a) Reference and
  body trajectory with camera locations c--e. (b) The robot starts within the
  arrival tolerance, leaves, and returns. (c--e) Original onboard frames from
  that recording. Departure, trajectory, and return records establish the
  navigation-level completion sequence.}
  \label{fig:real-loop}
\end{paperanchoredfigure}

\subsection{Supplementary Text-Driven Camera Task}

Separately from the voice-triggered inspection, we examine a supplementary
text-driven camera task. This case records a Qwen-Plus plan: capture,
persist evidence, and describe the image. All nodes
succeed without motion commands and share the image checksum, documenting
model-proposed task decomposition on real sensor data.

\subsection{Harness and RSI Evaluation Protocol}
\label{sec:extension-evaluation}

The assessment protocol separates task execution from cross-mission
configuration updates. Execution checks address evidence validity, completion
decisions, dispatch identity, resource ownership, and bounded recovery. RSI
checks address candidate changes, regression behavior, release decisions,
and rollback. Both use explicit acceptance criteria rather than a model's
unverified declaration of success.

For controlled harness comparisons, the protocol fixes the navigation
provider and routes across direct provider completion, monitored execution,
and evidence-gated execution. It specifies stale or missing pose evidence,
early provider success, repeated requests, and interrupted execution as fault
conditions. The ablation design removes evidence binding, role-isolated
contexts, or bounded recovery individually. Its outcome measures are
independently adjudicated task success, false completion, duplicate physical
dispatch, recovery outcome, and execution overhead.

For RSI, the protocol compares a frozen harness version with an accepted
candidate while holding the model and skill provider fixed. Each revision
retains the motivating traces, configuration difference, regression outcomes,
release decision, and rollback record. A held-out task set separates revision
generation from evaluation. Target measures are false completion, recovery
cost, regression failures, and release overhead. For VLM + \texttt{omni\_planner},
each exploration decision links the camera observation, proposed target,
planner response, and map update under one mission identity. These criteria
define how execution outcomes and configuration changes are assessed within
the model, agent, and RSI loops.

\section{Related Work}

\paragraph{Agent loops and harnesses.}
ReAct interleaves reasoning and action~\cite{yao2023react}. Anthropic's agent patterns
and LangChain's harness account describe feedback-driven tool use and orchestration
outside the model~\cite{anthropic2024agents,trivedy2026harness}; these are engineering
articles, not robot evaluations. Recent research makes this systems perspective
explicit: Gu studies context governance, trustworthy memory, and dynamic skill
routing~\cite{gu2026scaling}, while Code as Agent Harness surveys executable
interfaces, stateful mechanisms, and verification~\cite{ning2026codeharness}.
Self-Harness mines execution failures, proposes harness edits, and retains edits
after regression tests~\cite{zhang2026selfharness}. This motivates separating
runtime execution from cross-mission harness revision. Our RSI protocol
(Section~\ref{sec:rsi}) binds candidate revisions to robot execution evidence,
fixed acceptance checks, versioned release, and rollback. The agent loop retains
authority over physical task-state commitment under each accepted configuration.

\paragraph{Guarded execution.}
EmbodiedSkills is closely related: a fixed skill interface connects high-level
selection, preflight checks, bounded VLA execution, post-action verification, and
structured trajectories~\cite{wang2026embodiedskills}. CommitFlow also addresses
premature stage progression; it monitors physical requirements, blocks dependent
actions, and applies calibrated local corrections before
re-verification~\cite{zhao2026commitflow}. These works share our motivation and
preclude treating verification-gated progression itself as a new principle.
Our emphasis is the surrounding systems contract: C1--C4 information-access
boundaries, identity- and version-bound commit authority, duplicate-dispatch
control, resource ownership, and durable recovery accounting across heterogeneous
providers. These contracts define the mechanisms for a matched evaluation
with a fixed skill provider and common failure conditions.

\paragraph{World models and diagnosis.}
EV-WM~\cite{wang2026evwm} decodes imagined feature-space futures into task events and
uses event scores to guide candidate selection.  Onto-EV-WM~\cite{wang2026ontoevwm}
provides typed fact grounding and failure diagnosis around world-model prediction and
verification-gated correction.  Such methods can supply candidate assessments or
diagnostic information to a robot runtime.  \system{} addresses the execution contract
around heterogeneous providers: task-state commitment, role-specific contexts, skill
lifecycles, and bounded recovery.  Its runtime does not require either method, and
evaluating its contribution is distinct from evaluating prediction or correction quality.

\paragraph{Language-guided robot planning.}
Language models have been used to select robot skills by combining semantic likelihood with
learned affordances~\cite{ahn2022saycan}, and closed-loop language agents have incorporated
environment feedback into subsequent reasoning~\cite{huang2022inner}.
ReKep converts language and visual observations into relational keypoint constraints
and optimizes actions in a perception--action loop~\cite{huang2024rekep}.
These methods address skill selection, feedback, or geometric action construction.
\system{} instead specifies how a selected provider's execution is supervised and
how its evidence becomes accepted task progress; a planner's valid proposal is
not itself proof that the physical transition occurred.

\paragraph{Failure diagnosis and recovery.}
REFLECT summarizes multisensory execution history for language-based failure
explanation and correction~\cite{liu2023reflect}. AHA trains a vision-language
model to identify manipulation failures and explain them~\cite{duan2025aha}.
MAGMA-GEN generates recovery supervision by testing candidate corrections through
matched-state re-execution~\cite{bernat2026magmagen}. These target diagnostic
quality or the acquisition of recovery behavior, complementary to the verifier
and recovery-provider interfaces in our design.
Our C4 contract governs which recovery is authorized, how much budget it consumes,
and which new evidence must support subsequent acceptance. A useful diagnosis
and an authorized state transition remain different objects.

\paragraph{Evaluating recovery rather than nominal success.}
LIBERO-RECOVER evaluates manipulation from failure scenarios spanning action
retry, action adaptation, object-state recovery, and environmental
recovery~\cite{liu2026liberorecover}. This motivates measuring recovery separately
from nominal task completion. Our deployment cases concern navigation and
inspection; the benchmark addresses manipulation recovery.
For a harness comparison, the provider and failure conditions must be held fixed
while measuring false completion, duplicate dispatch, recovery expenditure, and
verified task outcomes. These criteria define the controlled evaluation protocol
in Section~\ref{sec:extension-evaluation}.

\paragraph{Visuomotor policies.}
RT-2~\cite{brohan2023rt2}, OpenVLA~\cite{kim2024openvla}, and $\pi_0$~\cite{black2024pi0}
demonstrate increasingly broad vision-language-conditioned control.
$\pi_{0.5}$ uses co-training across diverse data sources to improve
open-world generalization~\cite{physicalintelligence2025pi05}. \system{} does not
compete with these policies.  It treats each as a versioned, bounded backend whose outputs
must satisfy the same lifecycle, resource, evidence, and verification contracts as a
classical controller.

\paragraph{Embodied runtimes and memory.}
HROS~\cite{yan2026hros} supplies our prior quadruped inspection runtime, integrating
autonomy, voice, memory, and reporting. \system{} builds on this deployment
foundation and upgrades the agent-harness execution contract.
HoloAgent-0 organizes an embodied runtime, hierarchical 3D memory, and robot skills through
a monitored execution loop~\cite{zhou2026holoagent}. PhyAgentOS provides session-based
coordination, a state-as-a-file protocol, verification, epistemic memory, and a staged
evaluation process~\cite{liu2026phyagentos}. \system{} is complementary but places
its main technical emphasis on state-commit authority: observed versus committed facts,
role-isolated C1--C4 contexts, identity- and version-bound evidence packets, and event-sourced
budgeted recovery.  These mechanisms make false completion and correlated self-verification
explicitly testable harness properties.

\paragraph{Robot middleware and task execution.}
ROS~2 supplies distributed communication and execution infrastructure~\cite{macenski2022ros2},
behavior trees structure modular control flow~\cite{iovino2022behaviortrees}, and
TAMP integrates discrete task planning with continuous motion
planning~\cite{garrett2021tamp}. Lee et al. explicitly frame robot middleware as a
Physical AI harness, with output projection, execution/communication isolation,
and fallback transfer~\cite{lee2026physicalharness}. That enforcement layer is
complementary to our task-level contract: \system{} does not replace real-time
scheduling, network isolation, or low-level safety. It governs evidence provenance,
task-state acceptance, and recovery budgets above the provider interface.

\section{Conclusion}

\system{} organizes long-horizon execution around explicit authority and
evidence-bound state transitions. Role-isolated contexts, supervised skill
lifecycles, guarded commitment, and bounded recovery connect model proposals
to accepted task progress under the verifier and backend contracts. The
warehouse mission demonstrates an integrated physical workflow; the circuit
exposes history-dependent completion. The RSI loop applies the same
evidence-before-acceptance principle to harness revisions through regression
checks, versioned release, and rollback.

\bibliographystyle{plain}
% Bibliographic names and URLs leave few legal breakpoints. Do not stretch
% their short lines to the full column width or insert elastic block gaps.
\begingroup
\small
\raggedright
\renewcommand{\newblock}{\unskip\space}
\urlstyle{same}
\let\originalthebibliography\thebibliography
\renewcommand{\thebibliography}[1]{%
  \originalthebibliography{#1}\setlength{\itemsep}{2.5pt}}
\bibliography{references}
\endgroup

\end{document}